\documentclass[runningheads]{llncs}
\usepackage{fixltx2e} 
\usepackage{graphicx}
\usepackage{lipsum}
\graphicspath{{}}
\usepackage{caption}
\usepackage{subfig} 
\usepackage{amsmath}
\usepackage{amssymb}
\usepackage{epsfig}
\usepackage{hyperref}
\usepackage{todonotes}
\usepackage{footmisc}
\usepackage{textcomp}

\newcommand\blfootnote[1]{%
  \begingroup
  \renewcommand\thefootnote{}\footnote{#1}%
  \addtocounter{footnote}{-1}%
  \endgroup
}

\begin{document}
\title{An Advert Creation System for\\ 3D Product Placements}


\author{
Ivan~Bacher\thanks{Authors contributed equally.}\inst{1} \and
Hossein~Javidnia$^{\star}$\inst{1} \and
Soumyabrata~Dev$^{\star}$\inst{1,2} \and
Rahul~Agrahari\inst{1} \and 
Murhaf~Hossari\inst{1} \and 
Matthew~Nicholson\inst{1} \and 
Clare~Conran\inst{1} \and 
Jian~Tang\inst{3} \and  
Peng~Song\inst{3} \and  
David~Corrigan\inst{3} \and  
Fran\c{c}ois Piti\'e\inst{1,4}\thanks{The ADAPT Centre for Digital Content Technology is funded under the SFI Research Centres Programme (Grant 13/RC/2106) and is co-funded under the European Regional Development Fund.}
}
\institute{ADAPT SFI Research Centre, Trinity College Dublin, Dublin, Ireland \and
School of Computer Science, University College Dublin, Dublin, Ireland \and 
Huawei Ireland Research Center, Dublin, Ireland \and
Department of Electronic \& Electrical Engineering, Trinity College Dublin
}


\authorrunning{I. Bacher, H. Javidnia, S. Dev et al.}

\titlerunning{A 3D-Advert Creation System for Product Placements}
\maketitle 
\begin{abstract}
Over the past decade, the evolution of video-sharing platforms has attracted a significant amount of investments on contextual advertising. The common contextual advertising platforms utilize the information provided by users to integrate 2D visual ads into videos. The existing platforms face many technical challenges such as ad integration with respect to occluding objects and 3D ad placement. This paper presents a Video Advertisement Placement \& Integration (Adverts) framework, which is capable of perceiving the 3D geometry of the scene and camera motion to blend 3D virtual objects in videos and create the illusion of reality. The proposed framework contains several modules such as monocular depth estimation, object segmentation, background-foreground separation, alpha matting and camera tracking. Our experiments conducted using Adverts framework indicates the significant potential of this system in contextual ad integration, and pushing the limits of advertising industry using mixed reality technologies.
    \keywords{advertisement  \and augmented reality \and deep learning.}    
\end{abstract}
%
%
%
%
\section{Introduction}
\blfootnote{$^\dagger$Send correspondence to F.\ Piti\'e (PITIEF@tcd.ie).}
With the popularity of 4G networks and the decline in data traffic tariffs, the video content industry has maintained a relatively high growth rate. It is expected that the overall market size in 2021 will approach 211 billion RMB.\, an increase of 351\% as compared to 2018~\cite{china2020}. Such growing video demand and the increase of user generated videos creates additional challenges for advertisement and marketing agencies. The agencies need to devise innovative strategies to attract the attention of end-users. Traditionally, advertisements were added into existing videos as overlay, pre-roll, mid-roll or post-roll. These approaches are disruptive to the user's experience for online streaming applications.

In this paper, we solve the problem of disruptive user experience by creating a 3D-advertisement creation system. Therefore, this work describes a proof-of-concept prototype system that enables users to seamlessly insert a 3D object in any user-generated video. The user can select a 3D object from the library of 3D objects with in the proof-of-concept prototype, which can then be placed on any planar surface within a video scene. Our system can automatically analyze different depth layers in a video sequence and seamlessly integrate new 3D objects with proper occlusion handling.

\subsection{Related Work}
In the literature, there are several works in the area of advertisements in images and video streams. However, most of the existing work focus on the identification of logos and advertisement billboards in videos. Covell \emph{et al.} in \cite{covell2006advertisement} used audio and video features to accurately identify the sections in the video that contain the ads. This assists them to replace the existing ads with user-specific adverts in redistributed television materials. Hussain \emph{et al.} proposed a novel framework in \cite{hussain2017automatic} that understands the general sentiments of adverts using a large image- and video- datasets. Recently, Nautiyal \emph{et al.}\cite{nautiyal2018advert} used pre-trained deep learning models to identify existing 2D adverts in video streams, and seamlessly replace them with new adverts. Using large-scale annotated datasets of billboards~\cite{dev2019case,dev2019alos}, it provides them with an end-to-end framework for video editors to perform 2D advert placements. Their system assists in detecting frames in a video that contains a billboard~\cite{hossari2018adnet}, localizes the billboard in the detected frame~\cite{dev2019localizing}, and subsequently replace the existing billboard with a new 2D advertisement. In this paper, we generalize this problem of product placements into any user-generated videos, and artificially augment 3D adverts into the existing scenes.

\subsection{Contributions \& Organization of the paper}
Our contribution in this paper is two fold: (a) we propose a proof-of-concept prototype system that enables 3D computer graphic advertisement objects to be inserted seamlessly into video streams; and (b) we thereby establish a new paradigm in product placements for marketing agencies. Our proposed system will greatly assist video editors and content producers reduce the time it takes to dynamically generate augmented videos.

The remainder of the paper is arranged as follows: Section~\ref{sec:tech} briefly describe the technology behind our cloud-based advertisement creation system. Section~\ref{sec:sys-design} presents information regarding the design and development of the proof-of-concept prototype system. Section~\ref{sec:app} describes the various use cases and associated applications for the developed prototype. Section~\ref{sec:conc} concludes the paper. 


\section{Technology}
\label{sec:tech}
This section describes the methods and technologies employed in the Adverts framework including the monocular depth estimation module, camera tracking, interactive segmentation and background matting.

\subsection{Monocular Depth Estimation}
Monocular depth estimation is used to understand the 3D geometry of the scene and anchor the 3D plane on which the object will be placed. The classical depth estimation approaches heavily rely on multi-view geometry \cite{schonberger2016structure,dai2013projective,yu20143d,javidnia2017accurate,basha2012structure} such as stereo image \cite{scharstein2007learning,scharstein2002taxonomy}. These methods acquire depth information by utilising visual cues and different camera parameters which are not often available in offline monocular videos. The idea of using the monocular image to capture depth information could potentially solve the memory requirement issue of the conventional methods, but it is computationally difficult to capture the global properties of a scene such as texture variation or defocus information. The recent advancement of Convolutional Neural Networks (CNN) and publicly available datasets have significantly improved the performance of monocular depth estimation \cite{godard2017unsupervised,kuznietsov2017semi,bazrafkan2018semiparallel,fu2018deep,xu2018structured}. 

\begin{figure}[htb!]
\centering
\includegraphics[width=1\textwidth]{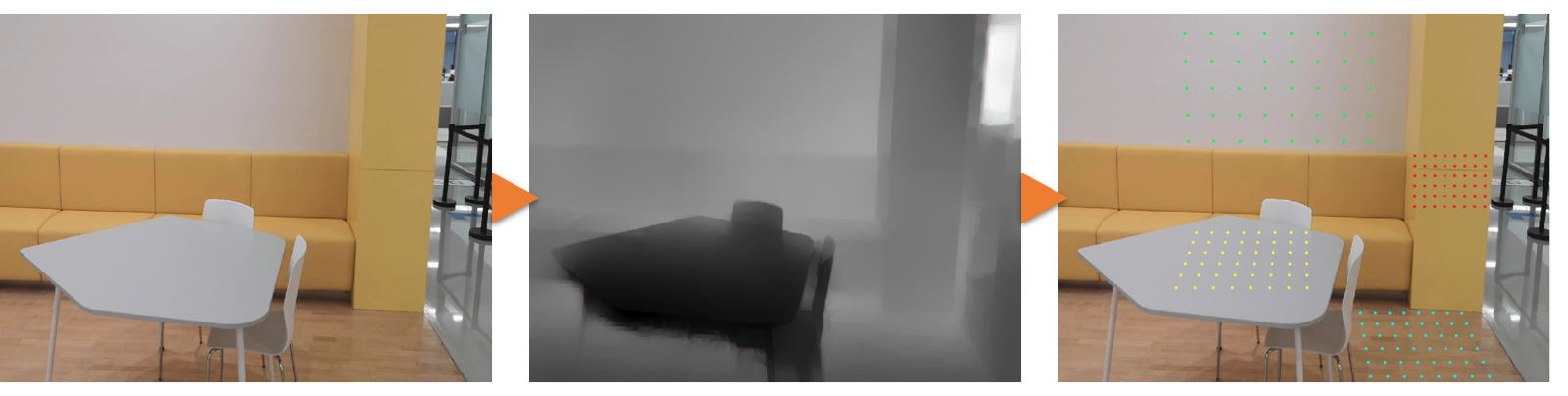}
\caption{Monocular depth estimation on a real-world scene. From left to right: input image; estimated depth map, localised plane using normal estimation.}
\label{fig:application-1}
\end{figure}

Several deep learning based monocular depth estimation networks are studied and evaluated in this research \cite{hu2019revisiting,li2018megadepth,lasinger2019towards,fu2018deep,tosi2019learning}. Among these, the network proposed by Hu \emph{et al.} \cite{hu2019revisiting} illustrated a superior performance in terms of accuracy and computational time compared to others. More importantly, this model showed a better generalization in depth scales due to the multi-scale feature fusion module integrated in the architecture. Figure \ref{fig:application-1} presents a sample of the monocular depth estimation followed by a localised plane in the scene. The orientation of the plane is obtained by calculating the normals from the depth information. The model by Hu \emph{et al.} \cite{hu2019revisiting} is employed as the first module in Adverts framework.

\subsection{Camera tracking}
\begin{figure}[htb!]
\centering
\includegraphics[width=1\textwidth]{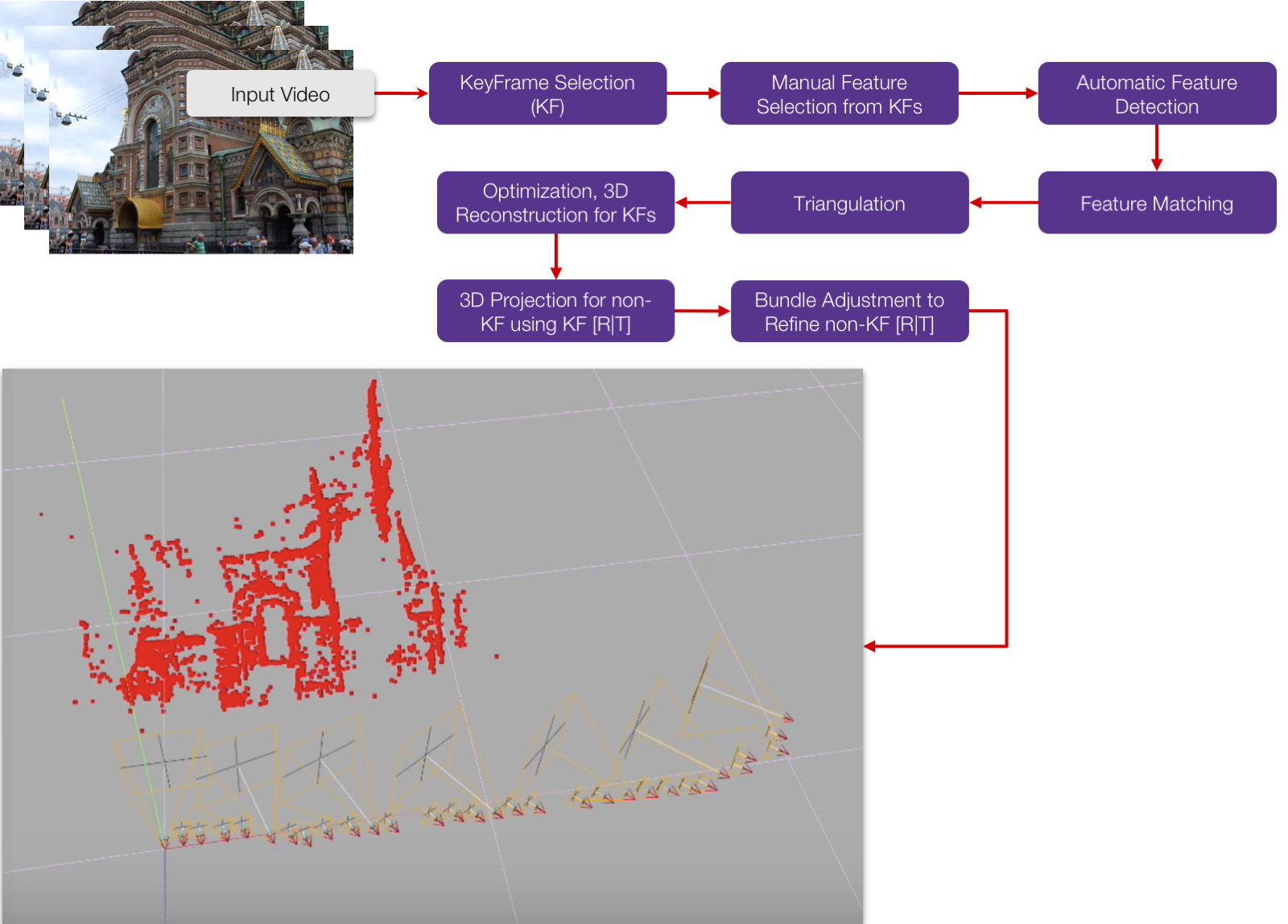}
\caption{Camera tracking pipeline implemented in Adverts framework.}
\label{fig:application-2}
\end{figure}

One of the very essential components of any augmented reality platform is tracking the camera motion to seamlessly integrate 3D object into the scene. Online augmented reality tools often utilise accelerometer, GPS, and solid state compass to track the camera motion in real-time. Such information is not available in offline scenarios. Adverts framework takes advantage of the traditional Structure from Motion (SfM) pipeline. Initially the user identifies a certain number of keyframes with manually matched feature points. Further, SIFT features \cite{lowe1999object} are detected and matched between the selected Keyframes followed by an optimization applied to refined the 3D projected points. The next step involves automatic feature matching between keyframes and non-keyframes. The matched features from  each non-keyframes are triangulated and reconstructed using the previous keyframes $[R|T]$. The final step of the camera tracking process involves a large scale sparse bundle adjustment \cite{agarwal2010bundle} with least square optimization applied to refine each non-keyframe\textquotesingle s $[R|T]$. Figure \ref{fig:application-2} illustrates the pipeline implemented to track the camera motion in Adverts framework. The camera projection matrix obtained for each frame is later used to project the 3D objects to camera space.

\subsection{Interactive Segmentation} \label{segmentation}
Determining the occluding object in an augmented reality application highly depends on the location of the 3D object and structural accuracy of the estimated depth map. By differentiating different layers of depth information, one can consistently integrate a virtual object in the scene. However, the quality of the integration result depends on the following measures:
\begin{enumerate}
  \item How accurate the general structure of depth map is?
  \item How temporally consistent the depth of the occluding object is across the entire video?
  \item How much detail is preserved in the depth structure of the occluding object?
\end{enumerate}

To compensate for the flaws of the estimated depth maps such as inaccurate depth boundaries, Adverts framework takes advantage of object segmentation in videos to produce binary masks of the occluding object. The generated masks are later used to estimate the alpha matte which is explained in the next section. This module allows users to interactively select the occluding object and decide which part of the scene is causing the occlusion by providing a broader control over tracking the occluding object across the entire video.

Similar to the depth estimation module, several methods are studied and evaluated for the segmentation part \cite{oh2019fast,jang2019interactive,maninis2018video,wug2018fast,wang2019fast,aksoy2018semantic,papazoglou2013fast,chen2018blazingly,lim2018learning,jain2017fusionseg}. The preliminary evaluation based on the DAVIS interactive segmentation benchmark \cite{caelles20182018} showed that the deep learning based model proposed by Oh \emph{et al.} \cite{oh2019fast} has a superior performance compared to the state of the art methods. This model was also ranked as the fastest one in the benchmark with the inference time of $0.2s$ for an image with $800\times600$ pixels resolution.
The network proposed by Oh \emph{et al.} \cite{oh2019fast} is constructed of two modules: interactive segmentation and mask propagation. The input to the interactive module is a tensor including a frame, the object mask from the previous round and two binary user annotation maps indicating the foreground and background regions. Further, the propagation module accepts a frame, its previous mask and the mask of the previous frame as the input to predict a new mask for the current frame. This model also utilises a Feature Aggregation Module designed to accumulate the information of the target object from all user interactions.

\begin{figure}[htb!]
\centering
\includegraphics[width=1\textwidth]{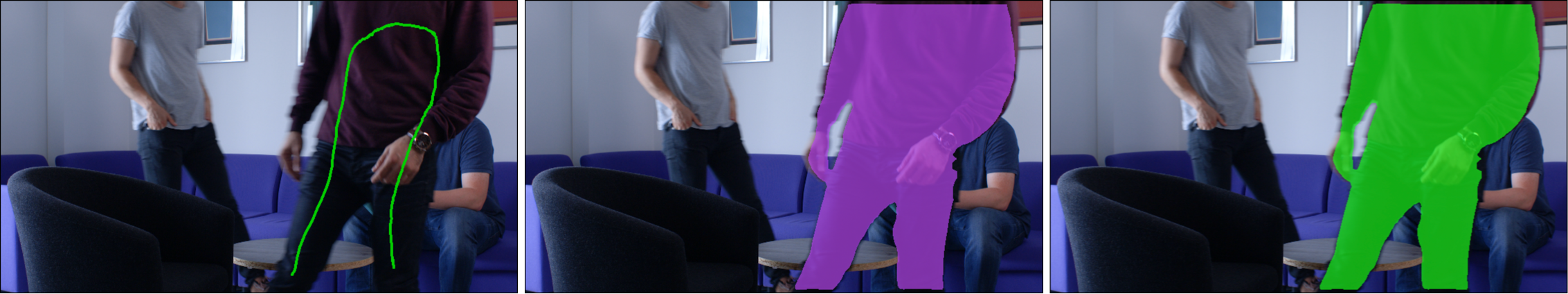}
\caption{Illustration of occlusion mask. User input, segmentation, propagation.}
\label{fig:application-3}
\end{figure}

The Adverts framework employs the model from \cite{oh2019fast} to interactively obtain the occlusion masks from users.
Figure \ref{fig:application-3} demonstrate an example of the interactive segmentation implemented in this paper.

\subsection{Background Matting}
To further refine the segmentation masks acquired from Section \ref{segmentation} and achieving fine level of details, the Adverts framework refers to alpha matte estimation. This is performed to calculate the opacity value of each blended pixel in the foreground object.

Generally, the composition image $I_i$ is represented as a linear combination of the background $B_i$ and foreground $F_i$ colors \cite{chuang2001bayesian}:
\begin{equation} \label{eq:1}
I_i = \alpha_iF_i+(1-\alpha_i)B_i
\end{equation}
where $\alpha_i \in [0,1]$ denotes the opacity or alpha matte of the foreground at pixel $i$. Often users provide guidance in a form of a trimap to solve this problem. Trimap assigns a label to every pixel as foreground $\alpha = 1$, background $\alpha = 0$ or unknown opacity. The goal of the alpha matting algorithms is to estimate the opacity value of the unknown regions by utilising the pixel values from known regions.
To achieve this goal, we investigated the effect of known background information in the matting process. This is done by introducing a Background-Aware Generative Adversarial Network to estimate alpha channels. Unlike the conventional methods, this architecture is designed to accept a 7 channel volume, where the first 3 channels contain the RGB image, the second 3 channels contain the RGB background information and the last channel contains the trimap. The preliminary experiments using the trained model indicates a significant improvement in the accuracy of the alpha mattes compared to the state of the art. The full details of this module including the background reconstruction and matting blocks are available as a preprint article on arXiv \cite{javidnia2020background}.

\section{System Design}
\label{sec:sys-design}
\begin{figure}[htb]
\centering
\includegraphics[width=1\textwidth]{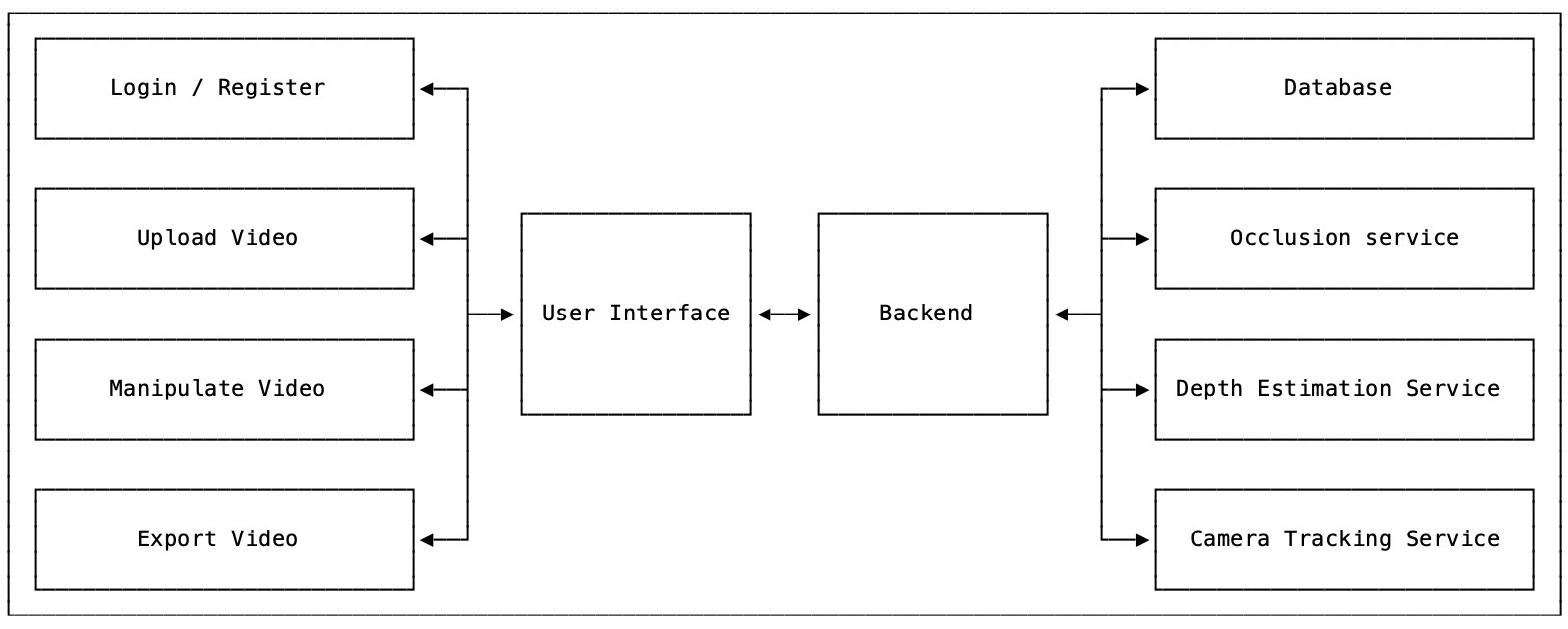}
\caption{System design of our proposed Adverts framework.}
\label{fig:4}
\end{figure}

In this section we describe the design and development, as well as the main technologies used for building the proof-of-concept prototype system that this work presents. The system can be split into two main components: user interface and back-end. Figure \ref{fig:4} illustrates the main structure of the system.

\subsection{User Interface}
\label{User Interface}
The user interface was implemented as a web application using modern web based technologies. This choice is supported by the fact that web based technologies only need a browser to run, thus making them cross platform compatible.
The main web based technologies utilised in the user interface are: 
Aurelia, 
Three.js,
Async,
Bootstrap, 
Fontawesome,
as well as state of the art web APIs such as 
the Broadcast Channel API, 
the Canvas API, 
the Fetch API,
and the Web Storage API\footnotemark[\value{footnote}].

\footnotetext{
\href{https://aurelia.io}{Aurelia}, 
\href{https://threejs.org/}{Three.js}, 
\href{http://caolan.github.io/async/v3/}{Async}, 
\href{https://getbootstrap.com/}{Bootstrap}, 
\href{https://fontawesome.com/}{Fontawesome}, 
\href{https://github.com/antimatter15/whammy}{Whammy}, 
\href{https://developer.mozilla.org/en-US/docs/Web/API/Broadcast_Channel_API}{Broadcast Channel API}, 
\href{https://developer.mozilla.org/en-US/docs/Web/API/Canvas_API}{Canvas API}, 
\href{https://developer.mozilla.org/en-US/docs/Web/API/Fetch_API}{Fetch API}, 
\href{https://developer.mozilla.org/en-US/docs/Web/API/Storage}{Storage API}
}

Aurelia can be described as a core building block of the user interface and is a JavaScript client framework for web, mobile, and desktops.
Three.js is an open source JavaScript library for 3D graphics on the web that supports WebGL. 
Async is a JavaScript library which provides functions for working with asynchronous code.
Bootstrap is an open source CSS framework and Fontawesome is an icon toolkit.
The Broadcast Channel API allows bi-directional communication between browsing contexts (windows, tabs, frames, or iframes) and workers on the same origin.
The Canvas API can be used for animation, game graphics, data visualisation, photo manipulation, and real-time video processing.
The Fetch API provides an interface for fetching resources.
The Web Storage API allows browsers to persistently store key/value pairs.

\begin{figure}[htb]
\centering
\includegraphics[width=1\textwidth]{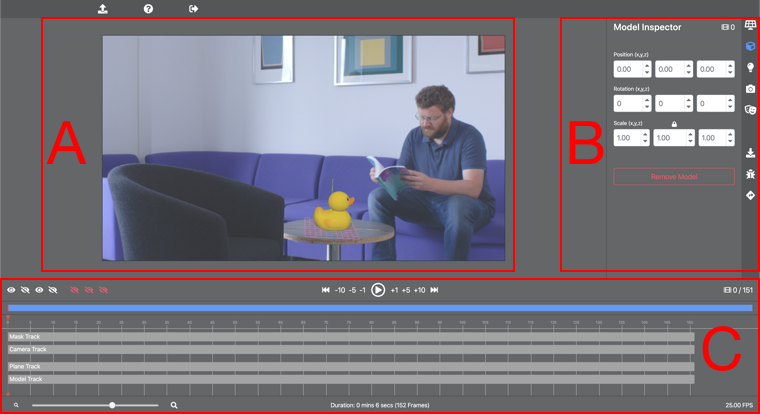}
\caption{User interface of our Adverts framework.}
\label{fig:5}
\end{figure}

\begin{figure}[htb]
\centering
\includegraphics[width=1\textwidth]{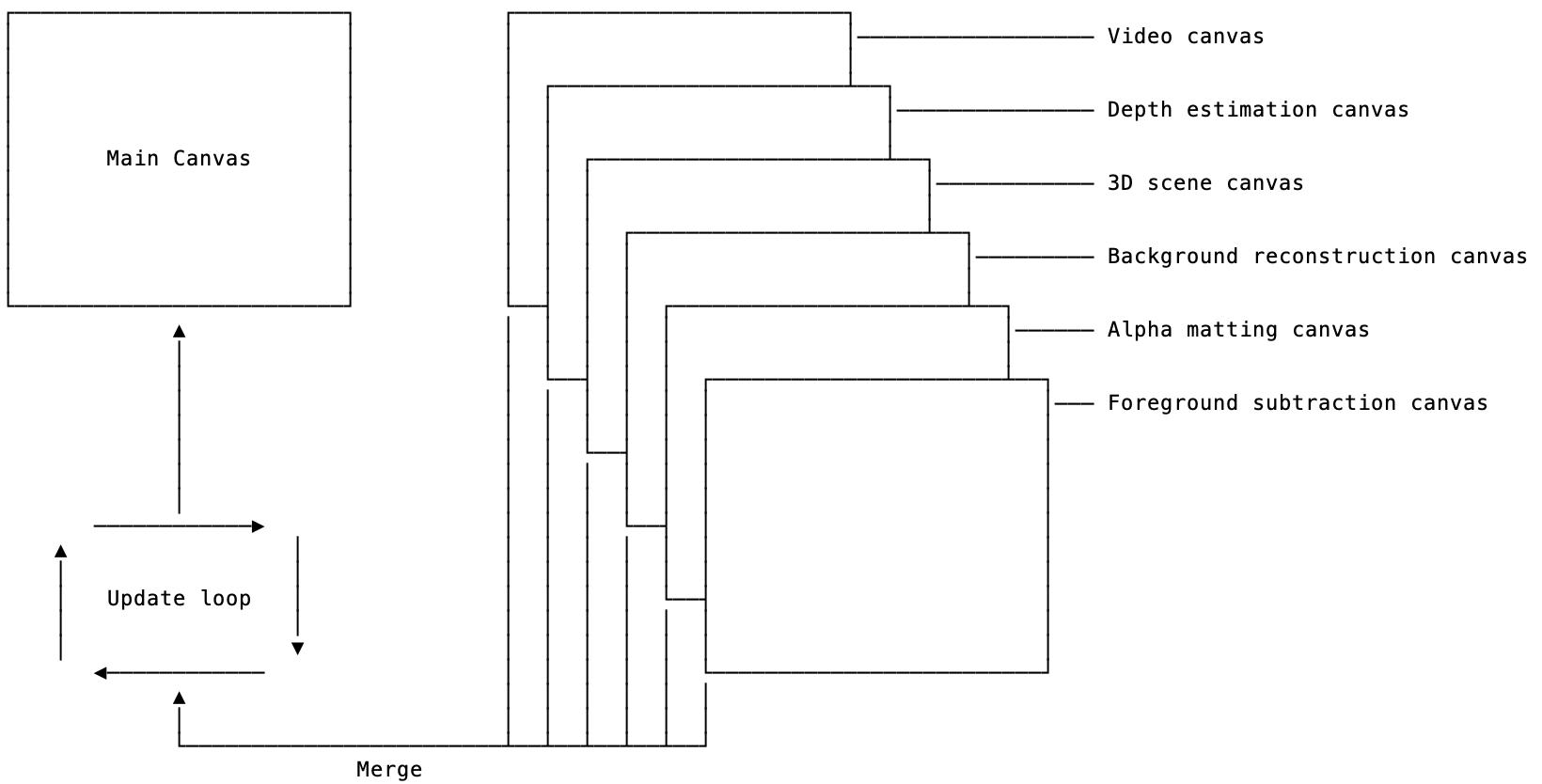}
\caption{Design of canvas layers in Adverts framework.}
\label{fig:6}
\end{figure}

Figure \ref{fig:5} depicts the video-editor of the user interface, which can be split into three main components.
Component A presents the user with a visual output of the manipulated video. It consists of a canvas element to which frames of the selected video, as well as 3D objects, occlusion masks, and depth estimation frames are drawn. If we examine Figure \ref{fig:5}, we can see that a 3D model of a yellow duck is superimposed onto the current selected video.
Figure \ref{fig:6} illustrates the design of the rendering canvas for component A of the video-editor page. The rendering canvas consists of one main canvas, and several hidden canvasses. The hidden canvasses are created in memory and each is responsible for rendering a specific video layer. The layers are composed of video frames of the selected video, depth estimation frames, 3D scene frames, background subtraction frames, alpha matting frames, and foreground reconstruction frames. The canvasses are updated ever iteration of the main rendering loop, where pixels are then extracted and merged into the main canvas.

Component B consists of a side-menu where the user is presented with a series of options. These options include opening the plane inspector, model inspector, light inspector, camera tracking inspector, mask inspector, as well as export and debug options. The plane inspector is used to add a plane to 3D scene superimposed  onto the current frame. The position of the frame is calculated using depth estimation information. This is crucial, as the plane acts as a anchor point for a 3D model. Users can use the plane inspector to change to positions, size, and rotation of the inserted plane. Further, the user also has the option to hide/show to plane.
The model inspector allows users to select one of the included 3D models, and add it to the plane within the superimposed 3D scene. Similar to the plane inspector, users can change the position, rotation, and size of the 3D model. Figure \ref{fig:5} shows the model inspector option currently selected. The light inspector allows users to manipulate lighting settings with the 3D scene to mimic those of the selected video. Users can switch lights on and off, as well as change the color, intensity, and position of various lights (ambient lights, spot lights, etc).
The camera tracking inspector allows users to add tracking points to the video on several key frames and start the tracking process in the back-end.
The mask inspector allows users to add an occlusion mask layer to the selected video for a range of frames. The user can select an object within the video that a mask should be created for.

Component C includes options that allow users to play, pause, fast forward, and reverse the selected video. Further, an overview timeline and a detailed timeline are also included in this component. The overview timeline shows the user in which frame they are currently located with the video. The detailed timeline shows the user which tracks have been added to the selected video. 

\subsection{Back-end}
\label{Back-end}
The development of back-end service is built upon the concept of Microservices Architecture. Reason for choosing this architecture lies behind its core concept of the Single Responsibility Principle which in simple terms means “gather together those things that change for the same reason and separate those things that change for different reasons.”

A microservices architecture takes this same approach and extends it to the loosely coupled services which can be developed, deployed, and maintained independently. Each of these services is responsible for a discrete task and can communicate with other services through simple APIs to solve a larger complex business problem. In our application, each module was independent of each other only those functionalities were grouped together which has a dependency with the previous module. Once the microservice is developed each one can be deployed independently which offer improved fault isolation whereby in the case of an error in one service the whole application doesn’t stop functioning. Another benefit which this architecture brings to the table is the freedom to choose technology stack(programming language, databases, cache etc) which is best suited for the service instead of using the one-size-fits-all approach.

\begin{figure}[htb]
\centering
\includegraphics[width=1\textwidth]{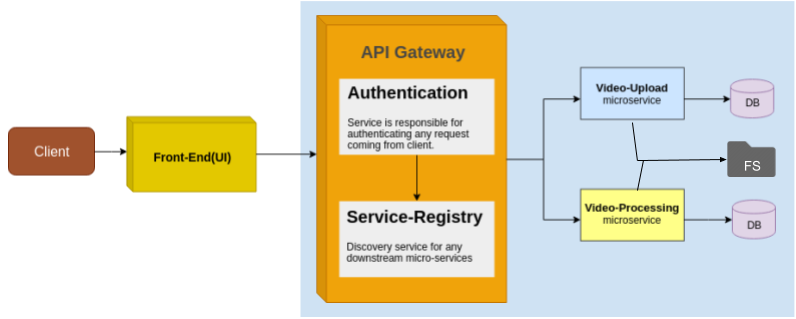}
\caption{Back-end Microservices Architecture in Adverts framework.}
\label{fig:7}
\end{figure}

The blue shaded region in Figure \ref{fig:7} is the back-end service where three major microservices work as a backbone to the whole application for providing core services. The three major services are:
API Gateway,
Video Upload,
Video Processing

\subsubsection{API Gateway}
\label{API Gateway}

This service is the first point of contact to any request coming from outside of the world. The API Gateway is responsible for authenticating the request and then navigating the request to the requested service. This service can also be called a proxy server as this is the interface for all the requests coming from the outside world and prevent direct exposure of the delicate services. The API Gateway accommodate the User Authentication and Service Registry Module

\paragraph{Authentication}
\label{Authentication}
For authenticating user request, JWT Authentication mechanism is being used which is one of the recommended standard authentication processes for microservices architecture. JWT is open industry-standard RFC 7519 for representing claims securely between two parties. It provides a compact and self-contained way for securely transmitting information between parties as a JSON object.  This information can be verified and trusted because it is digitally signed.[jwt.io].

\paragraph{Service Registry}
\label{Service Registry}

Service Registry is the discovery application for the microservices who want to use the API gateway for authentication and as a proxy server. We have used Eureka Service Registry developed and maintained by Netflix for their microservices which is robust and fast and known for its efficiency.

\subsubsection{Video Upload Service}
\label{Video Upload Service}
The input video upload is supported by video-upload services where the uploaded video is pre-processed i.e. the frames are extracted from the video and the frames are then stored in four different resolution (original, 480p, 720p, 1080p) along with storing the video property information in the database.


\subsubsection{Video Processing Service}
\label{Video Processing Service}

This is the service where the core functionalities are written and support the application. This service is majorly divided into three different components:Depth Estimation, Occlusion Detection, and Camera Tracking.


All these three components are dependent, hence they have been grouped together to support each other in functionality. Since the whole video processing is of heavy computation and time-consuming it does not make sense to restart the whole service from the beginning if the process is interrupted with an error at any stage. We go with marking the checkpoints at each stage and save the latest results of the processing. If the process is interrupted then the work will be resumed from the last checkpoint with loading the previous results whenever the process restarts.

\section{Application}
\label{sec:app}
\begin{figure}[htb]
\centering
\includegraphics[width=1\textwidth]{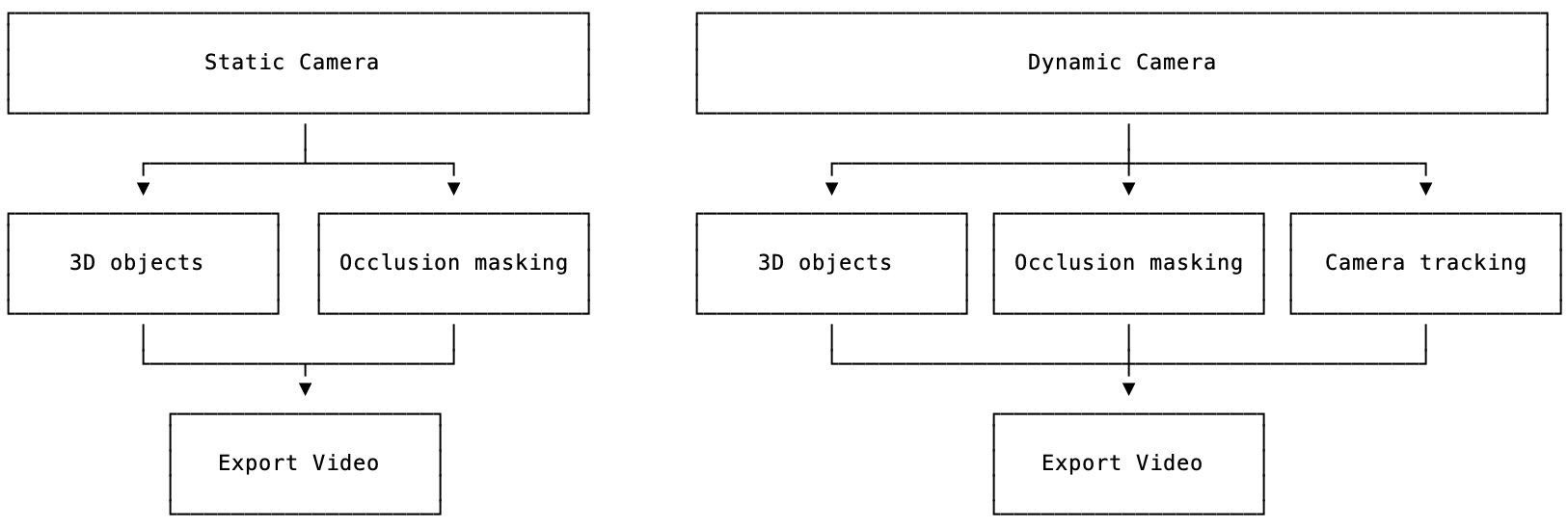}
\caption{Static and dynamic camera workflows.}
\label{fig:8}
\end{figure}

This section includes several examples of how the proof-of-concept prototype system can be used to dynamically generate augmented videos.
The examples include superimposing a 3D object into the existing video stream, creating an occlusion mask for the inserted 3D object, as well as tracking camera movement. Figure \ref{fig:8} illustrates these processes.

\subsection{Superimposing 3D Objects}
\begin{figure}[htb]
\centering
\includegraphics[width=1\textwidth]{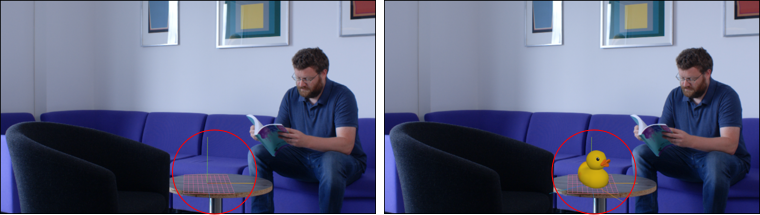}
\caption{Illustration of superimposing plane and 3D object onto video.}
\label{fig:9}
\end{figure}

Superimposing a 3D object onto a video stream consists of two main steps: 1) adding plane to the video and 2) adding a 3D model to the superimposed plane. Figure \ref{fig:9} shows an example of both steps.
To add a plane to the video, the user can select the plane inspector menu option (more details in section \ref{User Interface}). Once this option is selected, the user can move the mouse cursor over the video, which will display a temporary plane at the current position of the cursor. The user can then click the left mouse button to permanently add the plane to the video.
The position of the plane is calculated from the position of the mouse cursor, as well as from depth information obtained from the depth estimation module. Additionally, the user also has the option to manually change the position, rotation, and size of the superimposed plane, after it was added to the video.
To add a 3D object to the superimposed plane, the user can select the model inspector menu option (more details in section \ref{User Interface}). Once the menu option is selected, the user can choose a 3D object to add to the video from the 3D object library.
The user has the option to manually change the position, rotation, and scale of the 3D object.

\subsection{Occlusion masks}
\begin{figure}[htb]
\centering
\includegraphics[width=1\textwidth]{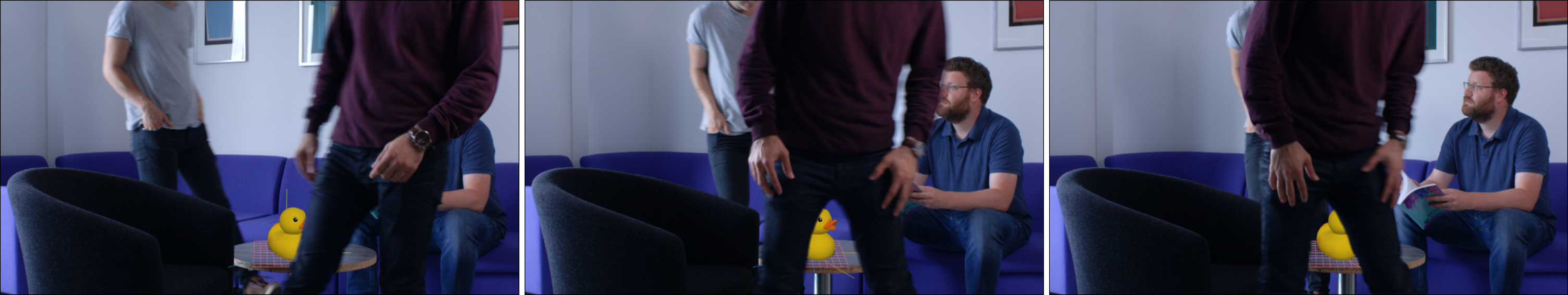}
\caption{Superimposing 3D object onto video with occlusion mask.}
\label{fig:10}
\end{figure}

To create an occlusion mask for a superimposed 3D object, the user must first determine the occluding area with the video. Once determined, the user can select this area and a segmentation mask is created using the interactive segmentation module described in Section \ref{segmentation}. The user can then choose to modify the segmentation mask, or propagate it over a range of video frames. After verifying if mask propagation was successful, the user can start the process of creating the occlusion masks for the range of selected frames. Once complete, the interface can be used to export the newly created video. Figure \ref{fig:10} depicts the the final result of adding a 3D object and an occlusion mask to the video.

\subsection{Camera Tracking}
\begin{figure}[htb!]
\centering
\includegraphics[width=1\textwidth]{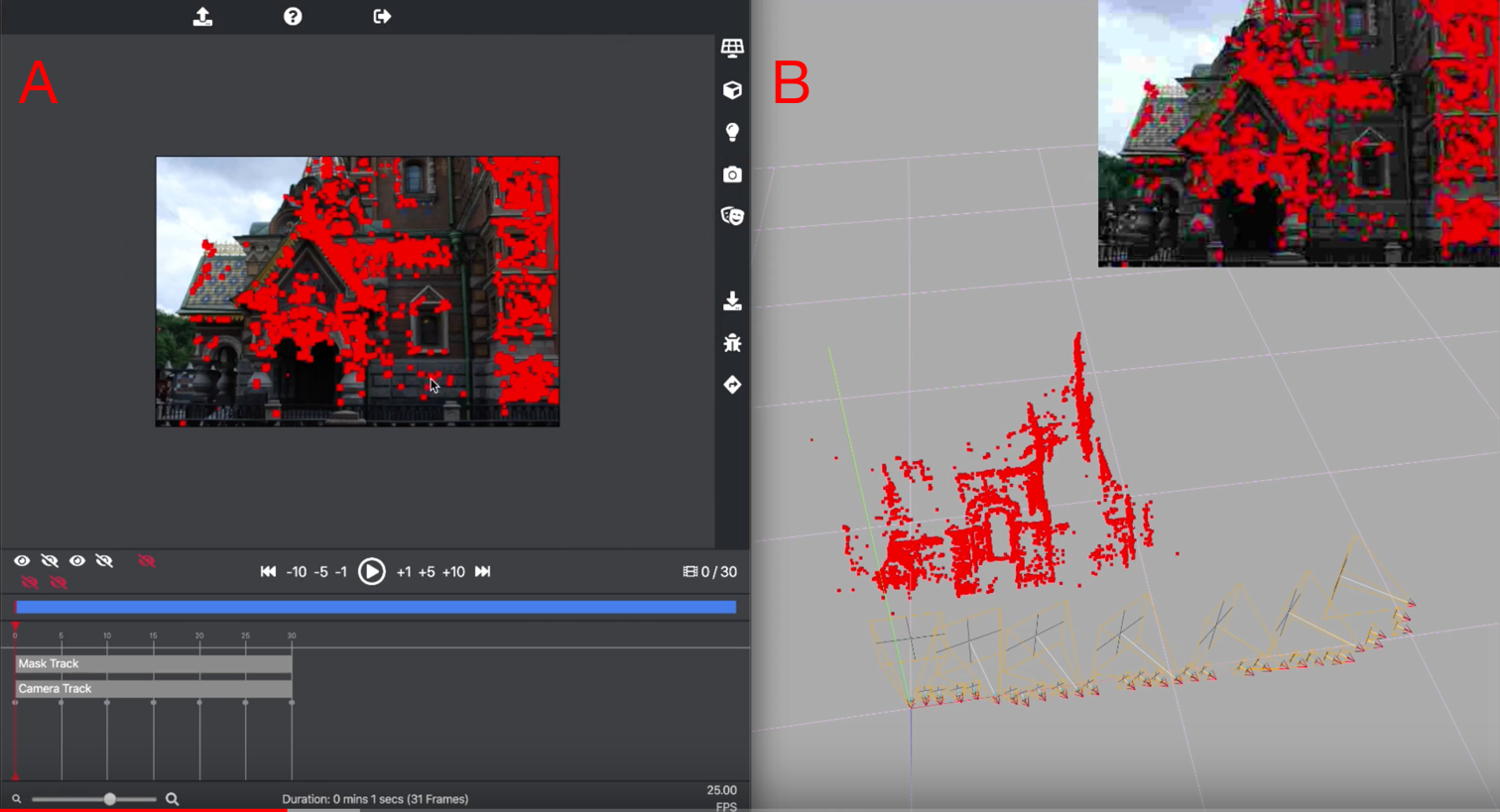}
\caption{Camera tracking - A) User interface B) 3D view of tracking information}
\label{fig:11}
\end{figure}

To track camera motion over a series of frames, the user must first manually add tracking feature points to the video. The user can add these points by using the mouse cursor to place a marker over the corresponding object within the video. Each frame that the user adds these points to becomes a keyframe. Once the user has completed adding the markers, the camera tracking process can be started. More information regarding this process can be found in Section \ref{sec:tech}. An additional window in the user interface is automatically opened, once the tracking process has completed. This window displays a 3D view of the video scene, in which matched feature points and debug cameras are drawn. Figure \ref{fig:11} B shows an example of this window. The red circles correspond to the matched feature points that the camera tracking algorithm returns. This additional window gives the user an overview of how the cameras position and orientation can change over a series of frames. Further, it can also be used to help with the placement of 3D objects and lights.

\section{Conclusion}
\label{sec:conc}

In this work we presented a proof-of-concept prototype  system  that  enables  3D  computer  graphic  advertisement  objects to be inserted seamlessly into video streams. Our goal was to  establish a new paradigm in product placements for marketing agencies and facilitate the process of dynamically generating augmented videos. We show that the proof of concept prototype can be used to inserted 3D objects into existing video streams, create occlusion masks for these objects, as well as track camera movement. 


%
%
\bibliographystyle{splncs04}

\begin{thebibliography}{10}
\providecommand{\url}[1]{\texttt{#1}}
\providecommand{\urlprefix}{URL }
\providecommand{\doi}[1]{https://doi.org/#1}

\bibitem{china2020}
China: short video market revenue 2016-2021 (Accessed 15-March-2020 2020),
  \url{https://www.statista.com/statistics/874562/china-short-video-market-size/}

\bibitem{agarwal2010bundle}
Agarwal, S., Snavely, N., Seitz, S.M., Szeliski, R.: Bundle adjustment in the
  large. In: European conference on computer vision. pp. 29--42. Springer
  (2010)

\bibitem{aksoy2018semantic}
Aksoy, Y., Oh, T.H., Paris, S., Pollefeys, M., Matusik, W.: Semantic soft
  segmentation. ACM Transactions on Graphics (TOG)  \textbf{37}(4),  1--13
  (2018)

\bibitem{basha2012structure}
Basha, T., Avidan, S., Hornung, A., Matusik, W.: Structure and motion from
  scene registration. In: 2012 IEEE Conference on Computer Vision and Pattern
  Recognition. pp. 1426--1433. IEEE (2012)

\bibitem{bazrafkan2018semiparallel}
Bazrafkan, S., Javidnia, H., Lemley, J., Corcoran, P.: Semiparallel deep neural
  network hybrid architecture: first application on depth from monocular
  camera. Journal of Electronic Imaging  \textbf{27}(4),  043041 (2018)

\bibitem{caelles20182018}
Caelles, S., Montes, A., Maninis, K.K., Chen, Y., Van~Gool, L., Perazzi, F.,
  Pont-Tuset, J.: The 2018 davis challenge on video object segmentation. arXiv
  preprint arXiv:1803.00557  (2018)

\bibitem{chen2018blazingly}
Chen, Y., Pont-Tuset, J., Montes, A., Van~Gool, L.: Blazingly fast video object
  segmentation with pixel-wise metric learning. In: Proceedings of the IEEE
  Conference on Computer Vision and Pattern Recognition. pp. 1189--1198 (2018)

\bibitem{covell2006advertisement}
Covell, M., Baluja, S., Fink, M.: Advertisement detection and replacement using
  acoustic and visual repetition. In: IEEE Workshop on Multimedia Signal
  Processing. pp. 461--466. IEEE (2006)

\bibitem{dai2013projective}
Dai, Y., Li, H., He, M.: Projective multiview structure and motion from
  element-wise factorization. IEEE transactions on pattern analysis and machine
  intelligence  \textbf{35}(9),  2238--2251 (2013)

\bibitem{dev2019alos}
Dev, S., Hossari, M., Nicholson, M., McCabe, K., Conran, C., Nautiyal, A.,
  Tang, J., Xu, W., Piti{\'e}, F.: The {ALOS} dataset for advert localization
  in outdoor scenes. In: 2019 Eleventh International Conference on Quality of
  Multimedia Experience (QoMEX). pp.~1--3. IEEE (2019)

\bibitem{dev2019case}
Dev, S., Hossari, M., Nicholson, M., McCabe, K., Nautiyal, A., Conran, C.,
  Tang, J., Xu, W., Piti{\'e}, F.: The {CASE} dataset of candidate spaces for
  advert implantation. In: 2019 16th International Conference on Machine Vision
  Applications (MVA). pp.~1--4. IEEE (2019)

\bibitem{dev2019localizing}
Dev, S., Hossari, M., Nicholson, M., McCabe, K., Nautiyal, A., Conran, C.,
  Tang, J., Xu, W., Piti{\'e}, F.: Localizing adverts in outdoor scenes. In:
  Proc. IEEE International Conference on Multimedia \& Expo Workshops (ICMEW).
  pp. 591--594. IEEE (2019)

\bibitem{fu2018deep}
Fu, H., Gong, M., Wang, C., Batmanghelich, K., Tao, D.: Deep ordinal regression
  network for monocular depth estimation. In: Proceedings of the IEEE
  Conference on Computer Vision and Pattern Recognition. pp. 2002--2011 (2018)

\bibitem{godard2017unsupervised}
Godard, C., Mac~Aodha, O., Brostow, G.J.: Unsupervised monocular depth
  estimation with left-right consistency. In: Proceedings of the IEEE
  Conference on Computer Vision and Pattern Recognition. pp. 270--279 (2017)

\bibitem{hossari2018adnet}
Hossari, M., Dev, S., Nicholson, M., McCabe, K., Nautiyal, A., Conran, C.,
  Tang, J., Xu, W., Piti{\'e}, F.: {ADNet}: A deep network for detecting
  adverts. arXiv preprint arXiv:1811.04115  (2018)

\bibitem{hu2019revisiting}
Hu, J., Ozay, M., Zhang, Y., Okatani, T.: Revisiting single image depth
  estimation: Toward higher resolution maps with accurate object boundaries.
  In: 2019 IEEE Winter Conference on Applications of Computer Vision (WACV).
  pp. 1043--1051. IEEE (2019)

\bibitem{hussain2017automatic}
Hussain, Z., Zhang, M., Zhang, X., Ye, K., Thomas, C., Agha, Z., Ong, N.,
  Kovashka, A.: Automatic understanding of image and video advertisements. In:
  Proceedings of the IEEE Conference on Computer Vision and Pattern
  Recognition. pp. 1705--1715 (2017)

\bibitem{jain2017fusionseg}
Jain, S.D., Xiong, B., Grauman, K.: Fusionseg: Learning to combine motion and
  appearance for fully automatic segmentation of generic objects in videos. In:
  2017 IEEE conference on computer vision and pattern recognition (CVPR). pp.
  2117--2126. IEEE (2017)

\bibitem{jang2019interactive}
Jang, W.D., Kim, C.S.: Interactive image segmentation via backpropagating
  refinement scheme. In: Proceedings of the IEEE Conference on Computer Vision
  and Pattern Recognition. pp. 5297--5306 (2019)

\bibitem{javidnia2017accurate}
Javidnia, H., Corcoran, P.: Accurate depth map estimation from small motions.
  In: Proceedings of the IEEE International Conference on Computer Vision
  Workshops. pp. 2453--2461 (2017)

\bibitem{javidnia2020background}
Javidnia, H., Piti{\'e}, F.: Background matting. arXiv preprint
  arXiv:2002.04433  (2020)

\bibitem{kuznietsov2017semi}
Kuznietsov, Y., Stuckler, J., Leibe, B.: Semi-supervised deep learning for
  monocular depth map prediction. In: Proceedings of the IEEE conference on
  computer vision and pattern recognition. pp. 6647--6655 (2017)

\bibitem{lasinger2019towards}
Lasinger, K., Ranftl, R., Schindler, K., Koltun, V.: Towards robust monocular
  depth estimation: Mixing datasets for zero-shot cross-dataset transfer. arXiv
  preprint arXiv:1907.01341  (2019)

\bibitem{li2018megadepth}
Li, Z., Snavely, N.: Megadepth: Learning single-view depth prediction from
  internet photos. In: Proceedings of the IEEE Conference on Computer Vision
  and Pattern Recognition. pp. 2041--2050 (2018)

\bibitem{lim2018learning}
Lim, L.A., Keles, H.Y.: Learning multi-scale features for foreground
  segmentation. Pattern Analysis and Applications pp. 1--12 (2018)

\bibitem{lowe1999object}
Lowe, D.G.: Object recognition from local scale-invariant features. In:
  Proceedings of the seventh IEEE international conference on computer vision.
  vol.~2, pp. 1150--1157. Ieee (1999)

\bibitem{maninis2018video}
Maninis, K.K., Caelles, S., Chen, Y., Pont-Tuset, J., Leal-Taix{\'e}, L.,
  Cremers, D., Van~Gool, L.: Video object segmentation without temporal
  information. IEEE transactions on pattern analysis and machine intelligence
  \textbf{41}(6),  1515--1530 (2018)

\bibitem{nautiyal2018advert}
Nautiyal, A., McCabe, K., Hossari, M., Dev, S., Nicholson, M., Conran, C.,
  McKibben, D., Tang, J., Xu, W., Piti{\'e}, F.: An advert creation system for
  next-gen publicity. In: Joint European Conference on Machine Learning and
  Knowledge Discovery in Databases. pp. 663--667. Springer (2018)

\bibitem{oh2019fast}
Oh, S.W., Lee, J.Y., Xu, N., Kim, S.J.: Fast user-guided video object
  segmentation by interaction-and-propagation networks. In: Proceedings of the
  IEEE Conference on Computer Vision and Pattern Recognition. pp. 5247--5256
  (2019)

\bibitem{papazoglou2013fast}
Papazoglou, A., Ferrari, V.: Fast object segmentation in unconstrained video.
  In: Proceedings of the IEEE International Conference on Computer Vision. pp.
  1777--1784 (2013)

\bibitem{scharstein2007learning}
Scharstein, D., Pal, C.: Learning conditional random fields for stereo. In:
  2007 IEEE Conference on Computer Vision and Pattern Recognition. pp.~1--8.
  IEEE (2007)

\bibitem{scharstein2002taxonomy}
Scharstein, D., Szeliski, R.: A taxonomy and evaluation of dense two-frame
  stereo correspondence algorithms. International journal of computer vision
  \textbf{47}(1-3),  7--42 (2002)

\bibitem{schonberger2016structure}
Schonberger, J.L., Frahm, J.M.: Structure-from-motion revisited. In:
  Proceedings of the IEEE Conference on Computer Vision and Pattern
  Recognition. pp. 4104--4113 (2016)

\bibitem{tosi2019learning}
Tosi, F., Aleotti, F., Poggi, M., Mattoccia, S.: Learning monocular depth
  estimation infusing traditional stereo knowledge. In: Proceedings of the IEEE
  Conference on Computer Vision and Pattern Recognition. pp. 9799--9809 (2019)

\bibitem{wang2019fast}
Wang, Q., Zhang, L., Bertinetto, L., Hu, W., Torr, P.H.: Fast online object
  tracking and segmentation: A unifying approach. In: Proceedings of the IEEE
  conference on computer vision and pattern recognition. pp. 1328--1338 (2019)

\bibitem{wug2018fast}
Wug~Oh, S., Lee, J.Y., Sunkavalli, K., Joo~Kim, S.: Fast video object
  segmentation by reference-guided mask propagation. In: Proceedings of the
  IEEE Conference on Computer Vision and Pattern Recognition. pp. 7376--7385
  (2018)

\bibitem{xu2018structured}
Xu, D., Wang, W., Tang, H., Liu, H., Sebe, N., Ricci, E.: Structured attention
  guided convolutional neural fields for monocular depth estimation. In:
  Proceedings of the IEEE Conference on Computer Vision and Pattern
  Recognition. pp. 3917--3925 (2018)

\bibitem{yu20143d}
Yu, F., Gallup, D.: 3d reconstruction from accidental motion. In: Proceedings
  of the IEEE Conference on Computer Vision and Pattern Recognition. pp.
  3986--3993 (2014)

\bibitem{chuang2001bayesian}
{Yung-Yu Chuang}, {Curless}, B., {Salesin}, D.H., {Szeliski}, R.: A bayesian
  approach to digital matting. In: 2001 IEEE Conference on Computer Vision and
  Pattern Recognition (CVPR). vol.~2, pp. 264--271 (Dec 2001).
  \doi{10.1109/CVPR.2001.990970}

\end{thebibliography}

\end{document}